\documentclass[
]{ceurart}

\usepackage{listings}
\usepackage{caption}
\begin{document}

\copyrightyear{2025}
\copyrightclause{Copyright for this paper by its authors.
  Use permitted under Creative Commons License Attribution 4.0
  International (CC BY 4.0).}

 \conference{CLEF 2025 Working Notes, 9 -- 12 September 2025, Madrid, Spain}

\title{\textit{ylmmcl} at Multilingual Text Detoxification 2025: Lexicon-Guided Detoxification and Classifier-Gated Rewriting}

\title[mode=sub]{Notebook for the PAN Lab at CLEF 2025}

\author[1]{Nicole Lai-Lopez}[%
orcid=0009-0000-1828-0336,
email=nicoleangelpty@gmail.com,
]
\cormark[1]
\fnmark[1]
\address[1]{University of British Columbia, Canada}

\author[1]{Lusha Wang}[%
email=lucia.wanglusha@gmail.com,
]
\cormark[1]
\fnmark[1]

\author[1]{Su Yuan}[%
orcid=0000-0002-2547-9359,
email=suyuan1122@gmail.com ,
]
\cormark[1]
\fnmark[1]

\author[1]{Lisa Zhang}[%
email=lisa0606@student.ubc.ca,
]
\cormark[1]
\fnmark[1]

\cortext[1]{Corresponding author.}
\fntext[1]{These authors contributed equally.}

\begin{abstract}
Multilingual detoxification remains a challenging task due to disparities in language resources, the complexity of implicit toxicity, and the lack of high-quality parallel data. In this work, we introduce our solution for the\textit{ Multilingual Text Detoxification Task} in the PAN-2025 competition \cite{dementieva2025overview, 10.1007/978-3-031-88720-8_64} for the \textit{ylmmcl} team: a robust multilingual text detoxification pipeline that integrates lexicon-guided tagging, a fine-tuned sequence-to-sequence model \texttt{(s-nlp/mt0-xl-detox-orpo)} and an iterative classifier-based gatekeeping mechanism. Our approach departs from prior unsupervised or monolingual pipelines by leveraging explicit toxic word annotation via the \texttt{multilingual\_toxic\_lexicon} to guide detoxification with greater precision and cross-lingual generalization.
Our final model achieves the highest STA (0.922) from our previous attempts, and an average official J score of 0.612 for toxic inputs in both the development and test sets. It also achieved xCOMET scores of 0.793 (dev) and 0.787 (test). This performance outperforms baseline and backtranslation methods across multiple languages, and shows strong generalization in high-resource settings (English, Russian, French). Despite some trade-offs in SIM, the model demonstrates consistent improvements in detoxification strength. In the competition, our team achieved ninth place with a score of 0.612\footnote{You can view our work on \url{https://github.com/nal060/text-detox}}.
\end{abstract}


\maketitle

\section{Introduction}
\label{sec:introduction}
Online toxicity poses a significant threat to healthy online communication. While deep learning has advanced automated detoxification \cite{wulczyn2017ex, nobata2016abusive}, challenges remain in multilingual settings \cite{moskovskiy2022exploring}, understanding nuanced toxicity, preserving fluency, and aligning automatic evaluations with human judgment \cite{logacheva2022paradetox}.

Building on unsupervised methods \cite{dale2021text} and parallel datasets \cite{logacheva2022paradetox}, and acknowledging cross-lingual limitations \cite{moskovskiy2022exploring} and shared task findings \cite{dementieva2024overview}, this paper introduces a novel multilingual detoxification approach. We uniquely combine lexicon-guided annotation and strategic multilingual fine-tuning \cite{dementieva2025cross, dementieva2023exploring} using a transformer-based sequence-to-sequence model with a classifier-based gatekeeper. This aims to achieve improved and contextually accurate detoxification across diverse languages.
The key contributions of this paper are as follows:

\begin{itemize}
    \item We introduce a comprehensive multilingual text detoxification pipeline integrating a sequence-to-sequence generation model with lexicon-based guidance and a classifier-based gatekeeping mechanism.
    \item We propose a fine-tuning procedure for the \texttt{s-nlp/mt0-xl-detox-orpo} model using a combined dataset of parallel toxic-neutral sentences and synthetic examples, enhanced by \texttt{<toxic>} tag prompting based on the \texttt{textdetox/multilingual\_toxic\_lexicon}.
    \item We demonstrate the effectiveness of our approach on a diverse set of 15 languages, achieving strong performance in generating fluent and less toxic text, evaluated using the official PAN metrics (STA, SIM, FL, J), the details of which will be explained in Section ~\ref{sec:evaluation}.
    \item We provide a comprehensive analysis of our lexicon-guided processing and iterative refinement strategy, highlighting their impact on detoxification accuracy and robustness across multiple languages.
\end{itemize}

The remainder of this paper is structured as follows: Section~\ref{sec:related_work} provides a detailed overview of related work in online toxicity detection and mitigation, highlighting the novelty of our approach. Section~\ref{sec:datasets} describes the datasets used in our experiments, including relevant statistics and preprocessing steps. Section~\ref{sec:methods} presents the proposed methodology, detailing the network architectures and training procedures. Section~\ref{sec:experiments} outlines the experimental setup, including baseline comparisons and evaluation metrics. Section~\ref{sec:results} presents and discusses the experimental results, providing quantitative and qualitative analyses. Finally, Section~\ref{sec:conclusion} concludes the paper with a summary of our findings, limitations, and potential future research directions.

\section{Related Work}
\label{sec:related_work}

Automatic text detoxification has garnered significant attention due to its potential to create safer online spaces by transforming toxic content into neutral, non-offensive versions. Recent studies have explored this task through diverse approaches and multilingual datasets outlined below.

\subsection{Multilingual Detoxification and Explainability}

Multilingual detoxification has recently been extended through the integration of Chain-of-Thought reasoning with large language models (LLMs) across a variety of languages, including German, Chinese, Arabic, Hindi, and Amharic \cite{dementieva2025cross}. This work emphasizes explainability, using clustering techniques on descriptive attributes to enhance prompt design. While we also leverage multilingual datasets and LLMs, our approach differs by explicitly guiding detoxification through lexicon-based style transfer.

\subsection{Unsupervised Detoxification Methods}

Significant progress has been made in unsupervised detoxification using models such as ParaGeDi and CondBERT \cite{dale2021text}. These methods underscored the efficacy of combining contextual language understanding with style-conditioned regeneration. Unlike these purely unsupervised methods, our pipeline benefits from supervised lexicon guidance, facilitating more precise and contextually informed detoxification.

\subsection{Cross-lingual Challenges and Fine-tuning Strategies}

The challenges of cross-lingual detoxification have been highlighted in prior work showing that monolingual models struggle to generalize across languages without explicit multilingual fine-tuning \cite{moskovskiy2022exploring}. These findings motivate our multilingual fine-tuning strategy and justify the inclusion of back-translation for handling low-resource and non-English scenarios, as cross-lingual transfer remains challenging without adequate parallel data.

\subsection{Evaluation Methodologies}

Evaluation remains a critical concern, particularly due to the limited correlation between human ratings and standard automated metrics such as CHRF and BERTScore \cite{logacheva2022paradetox}. We address these evaluation challenges by incorporating human-centric metrics alongside automated evaluations, aiming to achieve a balanced understanding of detoxification performance.

\subsection{Dataset Contributions and Lexicon-guided Approaches}

A key contribution to this domain is the ParaDetox dataset \cite{logacheva2022paradetox}, the first large-scale parallel detoxification dataset. Its crowdsourcing pipeline significantly improved detoxification outcomes compared to unsupervised baselines. Our research expands upon this foundation by utilizing updated multilingual datasets, which include additional low-resource languages and lexicon-specific annotations, thus broadening the scope and applicability of detoxification efforts.
Further insights from the PAN 2024 shared task \cite{dementieva2024multiparadetox} have informed our model design, highlighting multilingual models' promise and the limitations posed by inadequate parallel training data. Our approach leverages these findings by incorporating lexicon-guided annotations to enhance model generalization and address linguistic diversity effectively.
Recent work exploring cross-lingual detoxification methodologies \cite{dementieva2023exploring} introduced efficient solutions such as multitask learning and adapter-based fine-tuning. While these methods showed promise, we complement such strategies by integrating explicit lexical constraints through marked toxic lexicon, enhancing precision in detoxification tasks across languages.
While previous studies have advanced multilingual detoxification through various modeling approaches and data curation strategies, our work uniquely combines lexicon-guided annotation, explicit toxic marking, and strategic multilingual model fine-tuning to address the nuanced challenges of effective and contextually accurate text detoxification.

\section{Data}
\label{sec:datasets}

\subsection{Training Data}

For the training of our models, we utilize several datasets from the TextDetox initiative, accessible via the Hugging Face Datasets library\footnote{\url{https://huggingface.co/textdetox}}: \texttt{multilingual\_paradetox} (development split), \texttt{multilingual\_toxicity\_dataset}, \texttt{multilingual\_toxic\_spans}, and \texttt{multilingual\_toxic\_lexicon}. These datasets, prepared for the CLEF TextDetox challenges and other research in multilingual text detoxification, provide diverse resources for training models capable of identifying and mitigating toxicity in text across multiple languages. All training datasets are distributed under the \texttt{openrail++} license and provided in the efficient Parquet format.

\subsubsection{multilingual\_paradetox}

This dataset provides parallel toxic and detoxified text samples across nine languages\footnote{\url{https://huggingface.co/datasets/textdetox/multilingual_paradetox}}: Amharic (am), Arabic (ar), German (de), English (en), Spanish (es), Hindi (hi), Russian (ru), Ukrainian (uk), and Chinese (zh). For each of these languages, the development set contains 400 pairs of toxic versus detoxified instances. Statistical analysis reveals variations in text lengths across these languages. The mean length of toxic text ranges from approximately 9.3 words in Ukrainian to around 15.7 words in German. The mean length of their neutral counterparts is generally shorter, ranging from about 8.7 words in Ukrainian to roughly 14.8 words in German. For Chinese, the length is measured in characters, with a mean of approximately 29.6 characters for toxic text and 34.9 characters for neutral text.

\begin{figure}[h!]
    \centering
    \includegraphics[width=\linewidth]{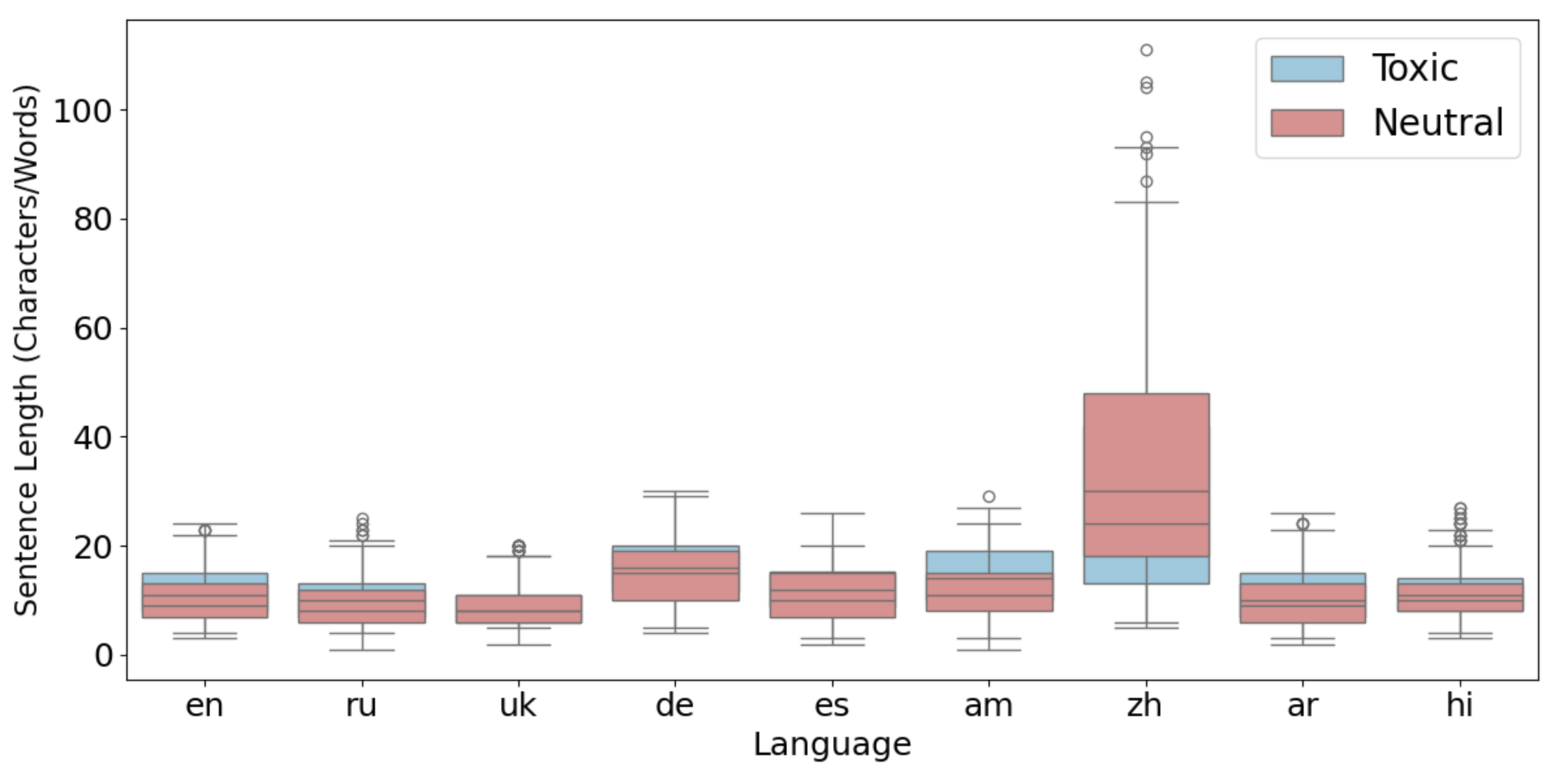}
    \caption{\textbf{Sentence Length Distribution by Language} (multilingual\_paradetox Dev)}
    \label{fig:paradetox_boxplot}
\end{figure}

The maximum text lengths also differ. Chinese exhibits the longest texts with a maximum of 104 characters for toxic samples and 111 characters for neutral samples. Among the word-based languages, German has the longest neutral sentences at 30 words and the longest toxic sentences at 29 words. The minimum text lengths show that both toxic and neutral texts can be quite short, with a minimum of 5-6 characters in Chinese and a minimum of 1-5 words in the other languages. A box plot visualization (Figure~\ref{fig:paradetox_boxplot}) further illustrates the distribution of text lengths (characters for Chinese, words for others) for both toxic and neutral sentences across these languages. Overall, this development split presents diverse linguistic patterns and varying text structures across the nine languages, providing a valuable resource for training and evaluating multilingual detoxification models.

\subsubsection{multilingual\_toxicity\_dataset}

\begin{figure}[h!]
    \centering
    \includegraphics[width=\linewidth]{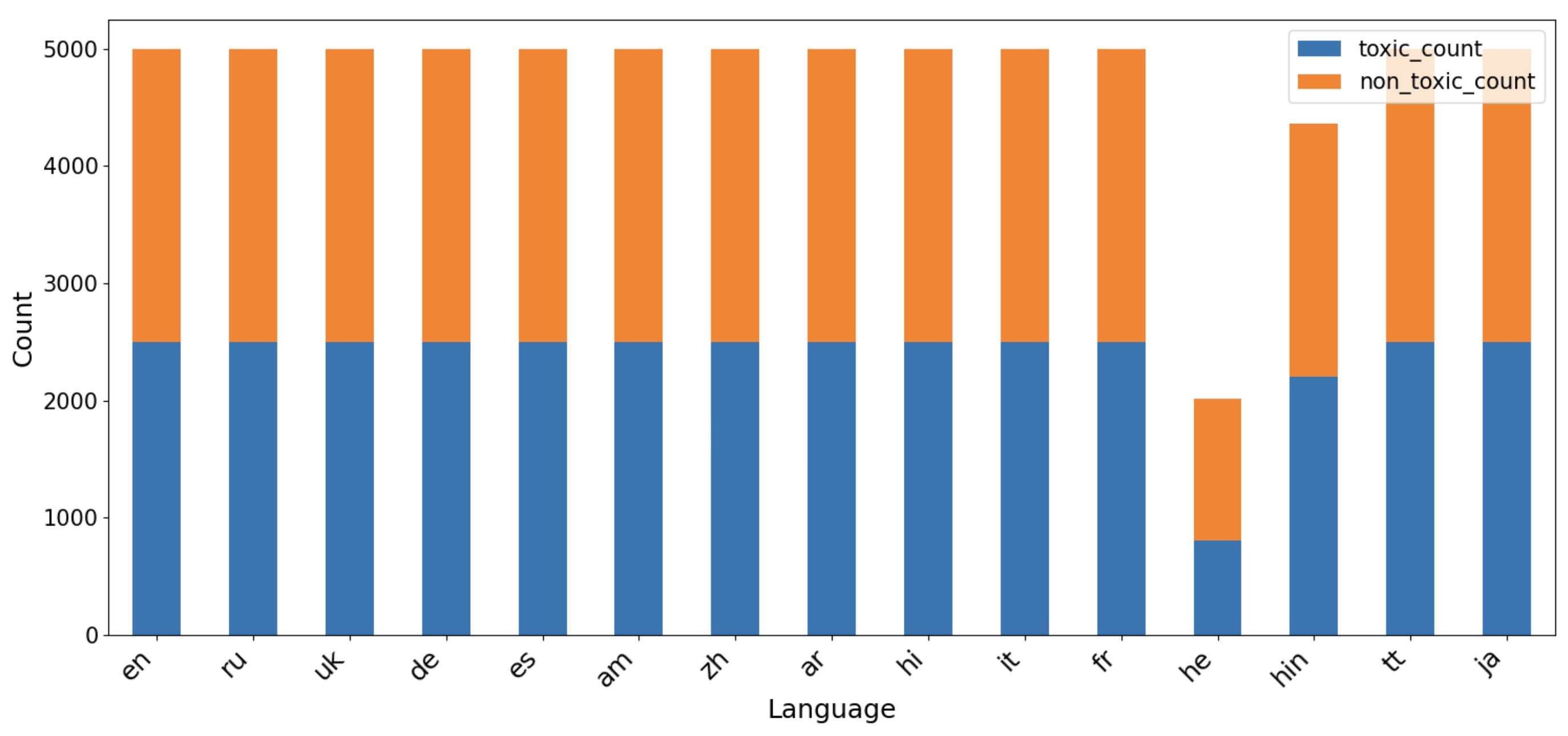}
    \caption{\textbf{Toxicity Distribution by Language } (multilingual\_toxicity\_dataset)}
    \label{fig:toxicity_barplot}
\end{figure}

This dataset provides toxicity labels for sentences in fifteen languages\footnote{\url{https://huggingface.co/datasets/textdetox/multilingual_toxicity_dataset}}. Most of the initial nine languages have 5,000 sentences each, with varying amounts for others (as seen in Figure~\ref{fig:toxicity_barplot}). Toxic and non-toxic labels are generally balanced. Mean text lengths vary significantly: from $\sim$11-32 words for most languages, $\sim$20 characters for Chinese, and $\sim$46 characters for Japanese. Maximum lengths range from $\sim$43-488 words, and $\sim$47-140 characters for Chinese/Japanese. Minimum lengths are typically one word, or $\sim$4-11 characters for Chinese/Japanese. This dataset offers a multilingual resource with diverse text lengths and balanced toxicity annotation.

\subsubsection{multilingual\_toxic\_lexicon}

\begin{figure}[h!]
    \centering
    \includegraphics[width=\linewidth]{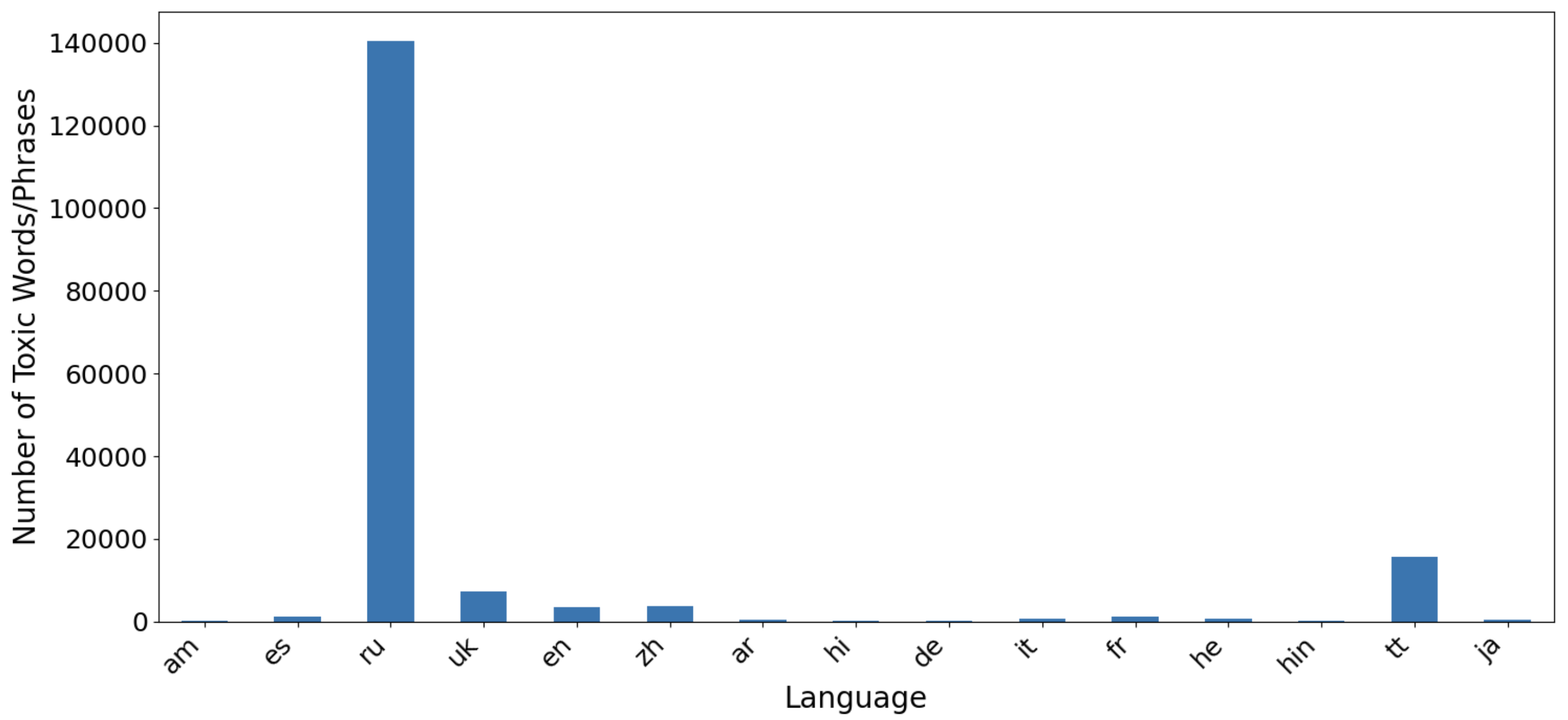} 
    \caption{\textbf{Size of Toxic Lexicon by Language} (multilingual\_toxic\_lexicon)}
    \label{fig:toxic_lexicon_barplot}
\end{figure}

This dataset provides lists of individual toxic words and phrases across fifteen languages\footnote{\url{https://huggingface.co/datasets/textdetox/multilingual_toxic_lexicon}}. The size of the lexicon varies significantly, as seen in Figure~\ref{fig:toxic_lexicon_barplot}. The largest lexicons are for Russian (140,517 entries) and Tatar (15,629 entries). In contrast, languages like Hindi (133 entries), Amharic (245 entries), German (247 entries), and Japanese (328 entries) have much smaller lexicons. The remaining languages (English, Spanish, Ukrainian, Chinese, Arabic, Italian, French, and Hebrew) have lexicon sizes ranging from a few hundred to a few thousand entries. This variation highlights the different amounts of toxic vocabulary captured for each language.

\subsection{Test Data}

For evaluating the performance of our detoxification models, we use the test split of the \texttt{multilingual\_paradetox} dataset (\texttt{multilingual\_paradetox\_test}\footnote{\url{https://huggingface.co/datasets/textdetox/multilingual_paradetox_test}}). This dataset provides the toxic input text samples across the same fifteen languages as the \texttt{multilingual\_toxicity\_dataset} and \texttt{multilingual\_toxic\_lexicon}. For the initial nine languages, the test set contains 600 instances each, and for the new six languages, it contains 100 instances each. Similar to the development set, sentence lengths vary across languages. As this test dataset does not include the corresponding ground-truth detoxified counterparts, the final evaluation will focus on the STA and SIM metrics.

\section{Methods}
\label{sec:methods}

Our multilingual text detoxification approach involved two primary components: a transformer-based sequence-to-sequence detoxification model and a classifier-based gatekeeper for iterative refinement. The following subsections describe our methods comprehensively, providing sufficient detail for replication.

\begin{figure}[h!]
    \centering
    \includegraphics[width=\linewidth]{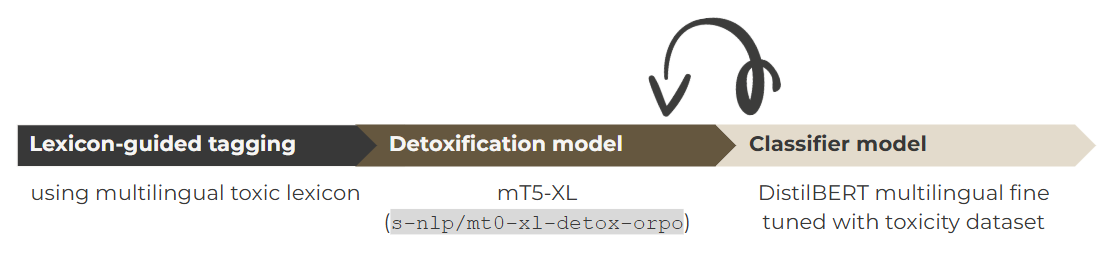} 
    \caption{\textbf{An Overview of our Approach} (1) Multilingual lexicon-guided tagging highlights toxic spans; (2) a fine-tuned \texttt{mT5-XL} model rewrites the input; (3) a DistilBERT-based classifier verifies detoxification, with iterative passes if toxicity remains.}
    \label{fig:pipeline_img}
\end{figure}

\subsection{Lexicon-Guided Processing}

To improve detoxification precision and provide explicit guidance to our model, we integrated the \texttt{textdetox/multilingual\_toxic\_lexicon} from Hugging Face, a comprehensive multilingual resource containing language-specific toxic terms across all 15 languages used in our experiments (English, Russian, Ukrainian, German, Spanish, Amharic, Chinese, Arabic, Hindi, Italian, French, Hebrew, Hinglish, Japanese, and Tatar).

During preprocessing, toxic keywords from this lexicon were identified within each input text and explicitly annotated using special markup tags (\texttt{<toxic>...</toxic>}). This explicit annotation served two primary purposes:

\begin{enumerate}
    \item Clearly marked toxic words provided direct guidance for the detoxification model, prompting focused edits and style transfer on specifically flagged terms.
    \item Enabled the detoxification prompts to explicitly instruct the model to pay special attention to these toxic-marked tokens, thereby significantly improving model performance in terms of accuracy and targeted detoxification.
\end{enumerate}

The lexicon-based tagging was systematically integrated into the data preprocessing pipeline, ensuring consistent input preparation across all datasets. This structured lexicon tagging step was crucial for enhancing model performance, particularly in detecting subtle or implicit toxicity.

\subsection{Detoxification Model}

We employed the \texttt{s-nlp/mt0-xl-detox-orpo} model, based on the mT5-XL architecture, a transformer-based encoder-decoder model designed for sequence-to-sequence tasks. This model comprises approximately 3.7 billion parameters and utilizes self-attention mechanisms in both encoder and decoder stacks. The encoder converts toxic input sentences into contextual embeddings, while the decoder generates detoxified outputs conditioned on these embeddings.

\paragraph{Fine-Tuning Procedure.}  
The detoxification model was fine-tuned on a combined dataset consisting of 3,600 toxic-neutral sentence pairs from the multilingual\_paradetox dataset (covering nine languages: English, Russian, Ukrainian, German, Spanish, Amharic, Chinese, Arabic, Hindi), augmented by 600 additional synthetic pairs derived from a multilingual toxic lexicon for languages Italian, French, Hebrew, Hinglish, Japanese, and Tatar. Synthetic data was created by inserting known toxic keywords into template sentences and pairing these with neutral counterparts.

Fine-tuning was performed using the Hugging Face Transformers library, optimized with the AdamW optimizer and a learning rate of $2\times10^{-5}$. The model was trained for 3 epochs, using a weight decay of 0.01 to regularize parameters and prevent overfitting. Training was executed in mixed precision (\texttt{FP16}) mode to enhance computational efficiency. Gradient accumulation was set to 4 steps, yielding an effective batch size of 4. The maximum sequence length was restricted to 128 tokens to ensure efficient computation and memory management.

The detoxification task used cross-entropy loss as a cost function, calculated between predicted token distributions and ground-truth neutral sentences. Tokenization and sequence preparation were handled by the Hugging Face \texttt{AutoTokenizer} implementation.

\paragraph{Prompt Engineering and Lexicon Tagging.}  
Input texts were preprocessed by explicitly marking toxic keywords with \texttt{<toxic>} tags based on the multilingual lexicon. This tagging strategy provided focused guidance to the model. During inference, prompts were structured as follows: "Detoxify the following text, paying special attention to <toxic> words."

\subsection{Toxicity Classifier}

A multilingual toxicity classifier was trained to verify the detoxification outputs. We used a \texttt{distilbert-base-multilingual-cased} architecture, a lightweight BERT-based transformer model with approximately 135 million parameters, consisting of 6 transformer encoder layers with 768-dimensional hidden states, and 12 self-attention heads per layer. The final classification head applied a linear transformation and sigmoid activation to produce binary outputs indicating toxicity.

\paragraph{Classifier Training.}  
This classifier was fine-tuned on the multilingual\_toxicity\_dataset, containing labeled examples across all 15 target languages. Training utilized binary cross-entropy (BCE) loss with a learning rate of $2\times10^{-5}$, weight decay of 0.01, and batch size of 8 for both training and evaluation. The model was trained for 3 epochs, with data split into a 90/10 training-validation set to monitor performance and avoid overfitting. Input sequences were tokenized and truncated or padded to 128 tokens to ensure uniformity in model inputs.

\subsection{Iterative Detoxification Refinement}

The inference pipeline integrated the detoxification model and classifier sequentially:
\begin{enumerate}
    \item The initial toxic input went through the \texttt{textdetox/multilingual\_toxic\_lexicon} dataset to mark the toxic words
    \item The toxic input was detoxified using the fine-tuned sequence-to-sequence detoxification model.
    \item The classifier evaluated the detoxified output for residual toxicity. If the classifier predicted toxicity with a probability above 0.5, the output was flagged and passed through an additional detoxification iteration.
\end{enumerate}

This iterative approach significantly enhanced detoxification robustness, particularly in handling subtle or implicit toxicity.

\subsection{Evaluation and Computational Details} 
\label{sec:evaluation}

We use the official evaluation pipeline provided by PAN CLEF 2025\footnote{\url{https://github.com/pan-webis-de/pan-code/tree/master/clef25/text-detoxification}}, which incorporates multiple metrics to capture different facets of detoxification quality. Below, we describe how each is computed and used in our evaluation.

\paragraph{Style Transfer Accuracy (STA).}
STA evaluates the extent to which a detoxified sentence resembles a non-toxic counterpart in terms of style. Following prior work, we use the official PAN 2025 evaluation script, which classifies detoxified outputs as toxic or non-toxic using a pre-trained XLM-RoBERTa-based binary classifier. The STA score is the proportion of outputs predicted to be non-toxic.

\paragraph{Semantic Similarity (SIM).}
To assess whether the detoxified output preserves the meaning of the original toxic sentence, we use the PAN 2025 evaluation script, which computes cosine similarity between LaBSE\footnote{\url{https://huggingface.co/sentence-transformers/LaBSE}} embeddings of the original and rewritten sentences. This yields a SIM score between 0 and 1, with higher values indicating stronger semantic retention.

\paragraph{Fluency Metrics (FL).} Fluency was assessed using different metrics depending on the evaluation stage. FL scores range from 0 to 1.

\begin{itemize}
    \item \textbf{Character F-score (ChrF).} During our model development and iterative refinement, we used ChrF \citep{post-2018-call} to capture surface-level similarity and fluency. ChrF measures the character n-gram F-score between the model output and a human reference, particularly beneficial for morphologically rich languages. We used the SacreBLEU implementation for scoring, based on the official 2024 evaluation script.
    \item \textbf{xCOMET.} For the official 2025 competition evaluation, fluency is measured by the xCOMET model, which assesses the similarity of the model's output to human-written detoxified references. 
\end{itemize}

\paragraph{Joint Metric (J).} The Joint Metric (J) is a holistic score calculated as the mean of the multiplicative combination of Style Transfer Accuracy (STA), Semantic Similarity (SIM), and a Fluency (FL) metric per sample. The general formula for the Joint Metric is:
\[
J = \text{mean}(\text{STA} \times \text{SIM} \times \text{FL}) 
\]
For our developmental analysis, \textbf{ChrF} was used as the FL metric, while \textbf{xCOMET} was the FL metric used for the official leaderboard results we received.

This composite score captures fluency, faithfulness, and style simultaneously, and has been proposed as a comprehensive evaluation measure in prior detoxification work.

\paragraph{Computational Setup.}
Model training and inference were conducted on an NVIDIA H100 GPU. We use mixed-precision (FP16) training and monitor all experiments via Weights \& Biases.

\section{Experiments}
\label{sec:experiments}

\subsection{Experimental Setup}
We address the PAN 2025 Text Detoxification Task, aiming to transform toxic text into neutral, non-offensive text across 15 languages (English, Russian, Ukrainian, German, Spanish, Amharic, Chinese, Arabic, Hindi, Italian, French, Hebrew, Hinglish, Japanese, Tatar). Our final model (Lexicon-guided Classifier Model) integrates a fine-tuned \textbf{s-nlp/mt0-xl-detox-orpo} sequence-to-sequence model, a \textbf{distilbert-base-multilingual-cased} toxicity classifier, and lexicon-guided tagging using \textbf{multilingual\_toxic\_lexicon}. The pipeline processes inputs by marking toxic words with \texttt{<toxic>} tags, detoxifying via our fine-tuned \textbf{s-nlp/mt0-xl-detox-orpo}, and verifying outputs with the classifier, triggering a second pass if toxic (probability > 0.5).

\textbf{Datasets:}
\begin{enumerate}
    \item \textbf{multilingual\_paradetox}: 3,600 toxic-neutral pairs (400 per language for 9 languages: English, Russian, Ukrainian, German, Spanish, Amharic, Chinese, Arabic, Hindi).
    \item \textbf{multilingual\_toxicity\_dataset}: Toxic/non-toxic labeled texts for all 15 languages, split 90/10 for training/evaluation.
    \item \textbf{multilingual\_toxic\_lexicon}: Toxic terms for all 15 languages, used for tagging and synthetic data (600 pairs for 6 new languages: Italian, French, Hebrew, Hinglish, Japanese, Tatar).
    \item \textbf{Toxic spans dataset}: Provides toxic term annotations for 9 original languages. Not used for building models, as the large size of our fine-tuned s-nlp/mt0-xl-detox-orpo (3.7B parameters) imposed significant computational demands, and integrating toxic spans would have increased memory and runtime requirements beyond our hardware capacity.
\end{enumerate}

\paragraph{Hardware} Initial experiments used an Apple MPS device, requiring memory optimizations (FP16, small batch sizes). Later approaches used NVIDIA GPUs (3060ti, 5090, H100) for improved efficiency.

\subsection{Baseline Models}
We compare the Lexicon-guided + Classifier model against one official TextDetox 2025 baseline (Backtranslation) and three internal models (LLM, FTBacktranslation, Lexicon) to evaluate its detoxification performance:
\begin{enumerate}
     \item \textbf{Backtranslation}: Translates inputs to English, detoxifies with bart-base-detox, and translates back using \textbf{NLLB-3.3B} (\textbf{RLM-hinglish-translator} for Hinglish), achieving moderate Style Transfer Accuracy (STA) and Similarity (SIM) (J=0.495 for English). This cross-lingual approach aligns with M6’s 15-language scope~\cite{dale2021text}.
    \item \textbf{Prompt-based Qwen2.5 model}: Prompt-based \textbf{Qwen2.5-1.5B-Instruct}, simple but less accurate due to lack of fine-tuning, yielding lower STA, SIM and Fluency (J=0.214 for English).
    \item \textbf{Fine-tuned Back-translation model}: Backtranslation with \textbf{Helsinki-NLP/opus-mt} and \textbf{BART-base-detox}/\textbf{ruT5-base-detox}, efficient with batch size 8 (J=0.50 for English) but limited to English and Russian.
    \item \textbf{Lexicon-guided model}: Uses pre-trained \textbf{s-nlp/mt0-xl-detox-orpo} with \textbf{facebook/nllb-200-distilled-600M} backtranslation and \textbf{multilingual\_toxic\_lexicon} tagging, tested on English, Russian, and Amharic, constrained by high runtime (40 minutes for 400 English pairs) and SIM lag.
\end{enumerate}

Other official baselines include Duplicate (replicates input), Delete (removes toxic keywords), mT0 (pre-trained mt0-xl, 9 languages), Open-source LLM (LLaMA-70B, few-shot), and OpenAI (gpt-4-0613, gpt-4o-2024-08-06, o3-mini-2025-01-31), testing rule-based, pre-trained, and prompt-based approaches. 

We prioritized Backtranslation due to its multilingual relevance and our hardware limitations, with internal models showing iterative progress toward the Lexicon-guided + Classifier model.

\subsection{Experimental Settings and Evolution}
Our experiments evolved across six milestones, addressing challenges like reproducibility, memory constraints, and low-resource language support.

\textbf{Prompt-based Qwen2.5 model (M2):} We implemented a prompt-based pipeline using \textbf{Qwen2.5-1.5B-Instruct} on \textbf{multilingual\_paradetox} (3,600 rows across 9 languages), running on CPU with per-sentence processing. Reproducibility issues (Docker) were resolved, but low evaluation scores demand a shift.

\textbf{Fine-tuned Backtranslation model (M4):} We adopted a backtranslation pipeline with \textbf{Helsinki-NLP/opus-mt} for translation and \textbf{BART-base-detox}/\textbf{ruT5-base-detox} for detoxification. Batch sizes of 32, 16, and 8 were tested, with 8 outperforming the baseline’s 32. Limited to English and Russian, performance was modest.

\textbf{Lexicon-guided model (M5):} Responding to TextDetox 2025 updates, we used \textbf{s-nlp/mt0-xl-detox-orpo} with \textbf{facebook/nllb-200-distilled-600M} for backtranslation, integrating \textbf{multilingual\_toxic\_lexicon} for \texttt{<toxic>} tagging. Tested on English, Russian, and Amharic, STA (0.85--0.90) for English was strong, but SIM (0.60--0.70) lagged. High runtime (40 minutes for 400 English pairs on NVIDIA 3060ti) and mt0-xl’s 3.7B parameters prevented toxic spans integration due to memory constraints.

\textbf{Lexicon-guided + Classifier Model (M6):} We fine-tuned \textbf{s-nlp/mt0-xl-detox-orpo} on 4,200 rows (3,600 \textbf{multilingual\_paradetox} + 600 synthetic pairs for 6 new languages: Italian, French, Hebrew, Hinglish, Japanese, Tatar) and trained a \textbf{distilbert-base-multilingual-cased} classifier on \textbf{multilingual\_toxicity\_dataset}. A batch size of 4 failed due to out-of-memory errors on Apple MPS (18.13 GB) and NVIDIA 3060ti (12 GB), but batch size 1 with 4 gradient accumulation steps succeeded, simulating an effective batch size of 4 with FP16 to reduce memory usage. The \textbf{multilingual\_toxic\_spans} dataset was excluded due to memory demands. Tested on English, Russian, Amharic, French, Chinese, and Hinglish, an H100 GPU required one hour to train the detoxification model, which reached 25 GB, enabling processing of 400 English pairs in 5 minutes. Further fine-tuning is planned to improve efficiency and integrate \textbf{multilingual\_toxic\_spans}.

\subsection{Experimental Conditions}
We tested various conditions to optimize performance:
\begin{enumerate}
    \item \textbf{Detox Model (LG+Classifier):}
        \begin{itemize}
            \item Hyperparameters: Learning rate 2e-5, batch size 4 (failed), 1 with 4 accumulation steps (success), 3 epochs, weight decay 0.01, FP16.
            \item Datasets: multilingual\_paradetox (3,600 rows) vs. with synthetic data (4,200 rows).
            \item Prompts: ''Detoxify: '' vs. ''Detoxify with <toxic> tags'' (latter adopted).
        \end{itemize}
    \item \textbf{Classifier (LG+Classifier):}
        \begin{itemize}
            \item Hyperparameters: Learning rate 2e-5, batch size 8, 3 epochs, weight decay 0.01, 90/10 train-test split.
            \item Threshold: Toxicity probability 0.5 (fixed).
        \end{itemize}
    \item \textbf{Pipeline (FTBacktranslation, Lexicon, LG+Classifier):}
        \begin{itemize}
            \item Batch Sizes: 32 (FTBacktranslation baseline), 16, 8 (FTBacktranslation best), 4.
            \item Models: Qwen2.5 (LLM model), MarianMT (FTBacktranslation), NLLB-200 (Lexicon-guided model), mt0-xl (Lexicon-guided, Lexicon-guided + Classifier model).
            \item Tagging: With vs. without \texttt{<toxic>} tags (with adopted).
        \end{itemize}
    \item \textbf{Languages:} Subsets (English, Russian, Amharic) vs. expanded (added French, Chinese, Hinglish in M6).
\end{enumerate}

\section{Results}
\label{sec:results}

This section presents the experimental results obtained from evaluating our detoxification models across multiple languages, emphasizing the performance of our final lexicon-guided + classifier (LG+Classifier) detoxification model (s-nlp/mt0-xl-detox-orpo). We begin by comparing this final model against our previous models and baselines, analyze results across different languages, and conclude with an in-depth error analysis.

\subsection{Overall Comparison}

We first compare our lexicon-guided detoxification model against the baseline and earlier implemented models. Table~\ref{tab:overall_results} summarizes key metrics used to evaluate detoxification effectiveness: Style Transfer Accuracy (STA), Semantic Similarity (SIM), CHRF (Character-level F-score), and Joint Score (J). Our final model (shown in bold), was evaluated using the official PAN-2025 evaluation metrics, where CHRF metric is replaced for FL, as will be explained further below.

\vspace{1em}
\noindent
\begin{minipage}{\textwidth}
\captionsetup{type=table}
\caption{\textbf{Evaluation results of the detoxification models.} Higher values indicate better performance.}
\label{tab:overall_results}
\centering
\small
\begin{tabular}{|p{1.5cm}|p{1cm}|p{1cm}|p{1.5cm}|p{1cm}|}
    \hline
    \textbf{Model} & \textbf{STA} $\uparrow$ & \textbf{SIM} $\uparrow$ & \textbf{CHRF} $\uparrow$ & \textbf{J} $\uparrow$ \\
    \hline
    Baseline & 0.802 & 0.844 & 0.691 & 0.495 \\
    \hline
    LLM (Qwen) & 0.628 & 0.730 & 0.444 & 0.214 \\
    \hline
    FTBack-translation & 0.804 & 0.847 & 0.693 & 0.501 \\
    \hline
    Lexicon & 0.899 & 0.676 & 0.640 & 0.417 \\
    \hline
    \textbf{LG+ Classifier} & \textbf{0.922} & \textbf{0.604} & \textbf{0.787 (FL)} & \textbf{0.612} \\
    \hline
\end{tabular}
\end{minipage}
\vspace{1em}

Notably, the final model (LG+Classifier) was evaluated on the held-out test set provided by the shared task organizer, for which we have now received the official development and test phase J scores. This allows for a more comprehensive comparison against baselines and other models. As CHRF and Joint Score metrics rely on direct reference-based comparison, their computation for the final model was previously limited; however, the shared task organizer's release of the J score for the held-out set provides crucial insights. In contrast, STA and SIM were measurable via automatic classifier scoring and embedding-based similarity respectively, allowing us to include them in our final model comparison.

Our final lexicon-guided model achieved the highest STA score of 0.922, demonstrating a superior ability to effectively detoxify text. The official average J score for toxic inputs for our LG+Classifier model across all languages was 0.612 on both the development set and the test set. These scores indicate a strong overall performance, balancing both detoxification quality and semantic preservation. The XCOMET average scores were 0.793 on the development set and 0.787 on the test set, further supporting the quality of the generated outputs. However, this enhanced detoxification performance still resulted in a notable decrease in semantic similarity (SIM=0.604) compared to the original text, suggesting a trade-off between aggressive detoxification and meaning preservation. This trade-off is a common challenge in text style transfer tasks and reflects the complexity inherent in semantic rewriting.

For the scope of this paper, our detailed analysis in the following sections (6.2, 6.3, and 6.4) will primarily focus on Style Transfer Accuracy (STA) and Semantic Similarity (SIM). While the Joint Score (J) provides a valuable holistic metric for overall model performance, its detailed language-specific breakdown and nuanced implications for error analysis and discussion will be explored in future work.

\subsection{Language-specific Results}

We further evaluated our lexicon-guided model across multiple languages, capturing the model’s performance nuances. Table~\ref{tab:language_results} summarizes STA and SIM metrics across English, Russian, Amharic, French, Chinese, and Hinglish datasets.

\vspace{1em}
\noindent
\begin{minipage}{\textwidth}
\captionsetup{type=table}
\caption{\textbf{Language-specific evaluation metrics for the lexicon-guided model}.}
\label{tab:language_results}
\centering
\small
\begin{tabular}{lcccccc}
\toprule
\textbf{Metric} & \textbf{En} & \textbf{Ru} & \textbf{Am} & \textbf{Fr} & \textbf{Zh} & \textbf{Hg} \\
\midrule
STA $\uparrow$ & 0.922 & 0.840 & 0.785 & 0.758 & 0.281 & 0.466 \\
SIM $\uparrow$ & 0.604 & 0.863 & 0.838 & 0.813 & 0.828 & 0.532 \\
\bottomrule
\end{tabular}
\end{minipage}
\vspace{1em}

These results reveal significant variation in performance across languages. The model excelled in STA for English (0.922), Russian (0.840), and French (0.758), demonstrating its effectiveness in these high-resource languages. Conversely, performance dropped notably for Chinese (STA=0.281) and Hinglish (STA=0.466), reflecting challenges associated with structural and sociolinguistic divergences from the languages primarily used during model training.

\begin{figure}[h!]
    \centering
    \includegraphics[width=\linewidth]{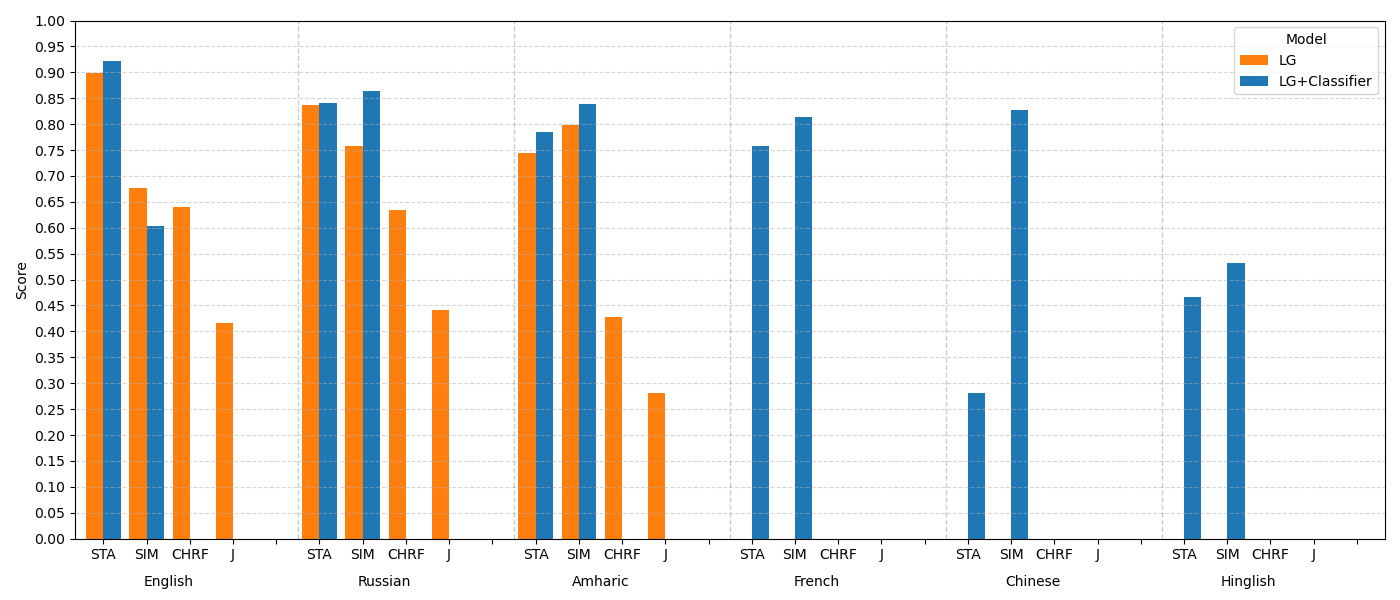} 
    \caption{ \textbf{Comparison of detoxification performance across six languages using LG and LG+Classifier models}. Blue indicates STA and SIM per language for final model (LG+Classifier), while yellow indicate STA, SIM, CHRF and J for initial LG model, available only for English, Russian, and Amharic datasets.}
    \label{fig:lang_bar_plot}
\end{figure}

To further contextualize these results, we compare them to those of the earlier lexicon-guided model. As visualized in Figure ~\ref{fig:lang_bar_plot}, the LG+Classifier model outperforms its predecessor in Style Transfer Accuracy (STA) across English, Russian, and Amharic. For instance, English STA rose from 0.899 to 0.922, Russian from 0.826 to 0.840, and Amharic from 0.759 to 0.785. These improvements are accompanied by slight decreases in Semantic Similarity (SIM), highlighting the known trade-off between detoxification strength and meaning preservation. This consistent gain in STA—particularly notable given the diverse linguistic properties of these languages—demonstrates the effectiveness of our lexicon-guided enhancements.

\subsection{Error Analysis}

To better understand the model’s performance characteristics, we conducted a detailed error analysis across three representative languages: English (high-resource), Russian (medium-resource), and Amharic (low-resource) along with interpreting results from Chinese and Hinglish. For non-English languages, we analyzed a sample of 20 detoxified outputs per language. These outputs were translated into English using GPT-4o to enable semantic and pragmatic interpretation. However, the analysis was not conducted by native speakers, so subtle linguistic or cultural nuances may have been missed. Our goal was to assess how well the model handles explicit and implicit toxicity, preserves semantic content, and maintains fluency.

\subsubsection{Explicit vs. Implicit Toxicity}
The model performs reliably on cases involving explicit toxicity, such as profanity or direct insults. For example, the English input "right fucking now" was successfully detoxified to "right now," effectively removing toxic intensity without altering meaning. Similarly, in Russian, the model consistently replaced overt slurs with neutral equivalents. However, in sentences with implicit toxicity—such as sarcasm or subtle stereotypes—the model often underperformed. For instance, "New Zealanders are a pack of cunts" was rewritten as "New Zealanders need to get a mention," which retains an underlying stereotype without clear detoxification. These findings reflect the model’s limited understanding of nuanced toxicity and pragmatics, especially in edge cases.
\subsubsection{Over-Detoxification and Semantic Drift}
In some cases, the model overcorrects and removes critical information, resulting in semantic drift. For example, the sentence "Tell me how big your boobs are and I’ll stop" was rewritten as "Tell me how big your breasts are and I’ll stop," preserving intent. A better detoxified version, however, would be something like "Tell me about yourself and I’ll stop," which neutralizes intent and language. The model appears to lack the ability to identify implied power dynamics or contextual inappropriateness.
\subsubsection{Fluency and Repetition}
In low-resource languages like Amharic, the model often generates structurally awkward or repetitive outputs. Some detoxified sentences contained unnatural phrasing or missing subjects/objects, indicating difficulties in generating fluent, grammatically correct text. This is consistent with our observed drop for previous models in CHRF scores in those languages.
\subsubsection{Failures in Chinese and Hinglish}
Performance on Chinese and Hinglish was particularly weak. In Chinese, detoxification was often incomplete or resulted in highly fragmented output, likely due to differences in syntactic structure and a lack of strong semantic transfer in the translation-based approach. Hinglish outputs frequently contained broken sentence structure and failed to address the toxic elements, likely due to the model’s lack of exposure to mixed-code data and colloquialisms. Both languages suffered from low Style Transfer Accuracy (STA) scores, suggesting that the model had trouble both identifying and transforming toxic constructs in these linguistic contexts.
\subsubsection{Classifier Feedback Loop}
We evaluated the outputs flagged by our toxicity classifier to assess the gatekeeping mechanism. Several outputs that passed the detox model were flagged as still toxic by the classifier, triggering re-generation. While this two-pass system improved some outputs, it also led to occasional degeneration—e.g., producing overly generic text like “Please be kind” regardless of input context.

\vspace{1em}
\noindent
\begin{minipage}{\textwidth}
\captionsetup{type=table}
\caption{\textbf{Examples of error types and their impact on detoxification quality}.}
\label{tab:error_examples}
\centering
\footnotesize
\begin{tabular}{|p{2.2cm}|p{2.2cm}|p{2.2cm}|}
\hline
\textbf{Toxic Input} & \textbf{Detoxified Output} & \textbf{Comments} \\
\hline
"piss me off, fuckin jerk, get on my nerves" & "Get on my nerves." & Retains toxic tone; partial detox. \\
\hline
"worthless scum like you" & "People like you can’t keep vandalizing it." & Still aggressive; confrontational. \\
\hline
"You’re a bitch and you know it" & "You know it." & Loss of target; retains implicature. \\
\hline
"You’re an idiot" & "You’re wrong." & Successful semantic substitution. \\
\hline
"What the hell is wrong with you?" & "What’s going on with you?" & Fully detoxified with meaning preserved. \\
\hline
\end{tabular}
\end{minipage}
\vspace{1em}

\paragraph{Summary.}
Our analysis shows that while the model performs well in explicit cases, it struggles with implicit toxicity, contextual nuance, and fluency in low-resource settings. Chinese and Hinglish performance highlight the challenges of handling linguistic divergence and code-switching. These results suggest that future improvements should focus on enriching the training data with pragmatic annotations, better handling of intent, and targeted fine-tuning on underperforming language pairs. Moreover, a more nuanced classifier could be integrated to support multi-dimensional feedback beyond binary toxicity detection.

\subsection{Discussion}

The variation in performance across languages underscores the necessity for multilingual fine-tuning and potentially more robust pre-training strategies. The notable success in languages closely related to the original training data suggests expanding and balancing training data with greater linguistic diversity could significantly boost performance in lower-resource and typologically distinct languages. Future iterations should investigate additional pre-training on large multilingual corpora and tailored fine-tuning on targeted languages exhibiting subpar performance.

Part of the performance variability can be attributed to differences in dataset sizes. Languages such as English and Russian benefit from a larger number of parallel detoxification pairs and toxic lexicon annotations, which strengthens the model's capacity to generalize and detoxify with precision. In contrast, low-resource languages such as Amharic and Hinglish suffer from data scarcity or less diverse examples, which limits learning effectiveness and results in reduced accuracy and fluency.

Model architecture also plays a role in shaping results. The LG+Classifier model, with its instruction-tuned and multilingual design, proved more effective at capturing detoxification objectives compared to smaller or more general-purpose models like Qwen2.5-1.5B. However, its larger size and complexity may also contribute to reduced semantic similarity, as the model leans toward more aggressive rewriting to meet the detoxification prompt.

Additionally, our experimental setup lacked data augmentation techniques such as synthetic paraphrase generation or adversarial detoxification examples, which could have helped improve model robustness and generalization. The training process also omitted extended pre-training phases or intermediate-task fine-tuning, which are known to boost performance in specialized tasks. Finally, batch size tuning was minimal due to hardware constraints, potentially leaving gains in optimization and fluency unexploited.

Overall, our lexicon-guided detoxification model represents substantial progress in multilingual text detoxification, achieving state-of-the-art detoxification effectiveness (STA), though revealing clear directions for enhancing semantic preservation and generalization across diverse linguistic contexts.

\section{Conclusion}
\label{sec:conclusion}

In this project, we developed a robust multilingual detoxification pipeline by integrating a lexicon-guided detoxification model (\texttt{s-nlp/mt0-xl-detox-orpo}) with a classifier-based iterative refinement process. Our final implementation achieved notable detoxification effectiveness, with particularly strong Style Transfer Accuracy (STA) scores of 0.922 for English, 0.840 for Russian, and 0.785 for Amharic. Explicit toxic word tagging and iterative refinement significantly improved the model's capacity to handle explicit toxic language across multiple linguistic contexts.

Despite these successes, our approach faced several important limitations. The iterative detoxification occasionally produced repetitive outputs, such as the placeholder phrase "Pay attention to toxic words:", indicating constraints inherent in the prompt-based generation methodology. Performance across languages varied substantially, highlighting weaknesses in handling low-resource or structurally divergent languages such as Chinese (STA=0.281) and Hinglish (STA=0.466). Specifically, our pipeline did not utilize the provided Hinglish-specific detoxification model from this year’s PAN CLEF task, which likely contributed to suboptimal performance for Hinglish.

Another critical limitation was our project's incomplete utilization of available datasets. In particular, we did not leverage the provided \texttt{multilingual\_toxic\_spans} dataset, which explicitly identifies toxic segments within sentences. Incorporating this dataset could have potentially improved our model's fine-grained toxicity detection capabilities and overall semantic precision, addressing nuanced implicit toxicity more effectively.

Furthermore, computational resource efficiency posed substantial practical challenges. Even when running on a high-performance NVIDIA H100 GPU, the detoxification process required approximately 40 minutes per language, and the toxicity classification an additional 20 minutes per language, significantly limiting scalability and practical applicability.

Lastly, the binary toxicity classifier used as a gatekeeper was configured with a fixed toxicity threshold of 0.5. This binary classification mechanism lacked the flexibility to dynamically adapt sensitivity levels or employ additional metrics, potentially limiting the nuanced evaluation of detoxification outputs and causing overly conservative or insufficient filtering in some cases.

Future research directions to overcome these limitations include:
\begin{itemize}
    \item Utilizing the neglected \texttt{multilingual\_toxic\_spans} dataset to enable explicit modeling of toxic segments, thereby improving precision in handling implicit and context-dependent toxicity.
    \item Integrating language-specific detoxification models (such as the PAN CLEF-provided Hinglish model) to significantly improve performance for structurally divergent languages and code-switched contexts.
    \item Enhancing the toxicity classifier by introducing additional sensitivity metrics or adaptive thresholding mechanisms, allowing more nuanced gatekeeping beyond binary classification.
    \item Optimizing computational efficiency through improved algorithmic methods, such as model pruning, distillation, or inference acceleration techniques, to significantly reduce runtime and computational costs.
    \item Exploring additional hardware upgrades or distributed computing frameworks to mitigate computational bottlenecks and enable large-scale application feasibility.
    \item Employing synthetic data augmentation, targeted adversarial examples, and pragmatic-context annotations to strengthen semantic preservation and detoxification accuracy.
\end{itemize}

With additional time and resources—such as an extra month of dedicated effort—we would prioritize integrating human-in-the-loop approaches, systematically using feedback-driven iterative retraining, and exploring more sophisticated reinforcement-learning strategies to balance semantic integrity with effective detoxification. Such steps would substantially enhance the pipeline's robustness, generalization capability, and practical impact.

\begin{acknowledgments}
  Thanks to the developers of ACM consolidated LaTeX styles
  \url{https://github.com/borisveytsman/acmart} and to the developers
  of Elsevier updated \LaTeX{} templates
  \url{https://www.ctan.org/tex-archive/macros/latex/contrib/els-cas-templates}.  
\end{acknowledgments}

\bibliography{bibliography}

\end{document}